%% file: main.tex
\documentclass[letterpaper, 10 pt, conference]{ieeeconf}  

\usepackage{graphicx}
\usepackage[ruled, vlined]{algorithm2e}
\usepackage{amsfonts}
\usepackage{amsmath}
\usepackage{multirow}
\usepackage{cite}
\usepackage{adjustbox}
\usepackage{mathtools}

\usepackage{hyperref}
\usepackage[T1]{fontenc}
\usepackage[utf8]{inputenc}
\usepackage[caption=false]{subfig}
\usepackage{booktabs}
\usepackage[dvipsnames]{xcolor}

\usepackage{pifont}
\newcommand{\cmark}{\ding{51}}%
\newcommand{\xmark}{\ding{55}}%

\makeatletter
\newcommand{\removelatexerror}{\let\@latex@error\@gobble}
\makeatother

\IEEEoverridecommandlockouts                              

\title{\LARGE \bf
Learning to Play Soccer From Scratch: Sample-Efficient Emergent Coordination through Curriculum-Learning and Competition
}

\author{Pavan Samtani$^{1}$ \and Francisco Leiva$^{2}$ \and Javier Ruiz-del-Solar$^{1, 2}$
\thanks{This work was supported by FONDECYT project 1201170, ANID-PIA
project AFB180004, and CONICYT-PFCHA/Mag\'ister Nacional/2018-22182130.}
\thanks{$^{1}$Department of Electrical Engineering, Universidad de Chile, Tupper 2001, Santiago, Chile.}%
\thanks{$^{2}$Advanced Mining Technology Center (AMTC), Universidad de Chile, Tupper 2007, Santiago, Chile.\newline{\tt\footnotesize \{pavan.samtani, francisco.leiva, jruizd\}@ing.uchile.cl}}%
\thanks{This work has been submitted to the IEEE for possible publication. Copyright may be transferred without notice, after which this version may no longer be accessible.}%
}

\begin{document}

\maketitle
\thispagestyle{empty}
\pagestyle{empty}

\begin{abstract}
This work proposes a scheme that allows learning complex multi-agent behaviors in a sample efficient manner, applied to 2v2 soccer.
The problem is formulated as a Markov game, and solved using deep reinforcement learning. We propose a basic multi-agent extension of TD3 for learning the policy of each player, in a decentralized manner. To ease learning, the task of 2v2 soccer is divided in three stages: 1v0, 1v1 and 2v2. The process of learning in multi-agent stages (1v1 and 2v2) uses agents trained on a previous stage as fixed opponents. In addition, we propose using experience sharing, a method that shares experience from a fixed opponent, trained in a previous stage, for training the agent currently learning, and a form of frame-skipping, to raise performance significantly. Our results show that high quality soccer play can be obtained with our approach in just under 40M interactions. A summarized video of the resulting game play can be found in \url{https://youtu.be/f25l1j1U9RM}.

\end{abstract}

\section{INTRODUCTION}
\label{sec:introduction}

\input{introduction}

\section{RELATED WORK}
\label{sec:related_work}
\input{related_work}

\section{PROPOSED APPROACH}
\label{sec:methodology}
\input{methodology}

\section{EXPERIMENTAL RESULTS}
\label{sec:results}
\input{results}

\section{CONCLUSION}
\label{sec:conclusion}
\input{conclusion.tex}


\bibliographystyle{ieeetr}
\bibliography{IEEEabrv, references}


\end{document}

%% file: introduction.tex
Multi-agent problems are especially challenging when coordination and competition between agents is encouraged and/or required. An instance of such problems is soccer, where agents of a given team collaborate to score goals against an opposing team.

Getting a team of autonomous agents to play soccer has been an open research problem for a long time. Some efforts to address this problem include several leagues of the RoboCup competition which foster research in the topic, as well as standalone simulated environments and benchmarks that have been proposed and open-sourced (e.g. \cite{liu2018emergent, kurach2020google}).

Although hand-crafted behaviors may endow a team with the ability to play soccer collaboratively, an important research question is whether or not such behaviors may be learned. In this regard, an increasingly popular approach for multi-agent learning corresponds to reinforcement learning (RL). Using RL, attempts have been made to address the problem of playing soccer and related sub-tasks (e.g. the ``\textit{keepaway}'' \cite{stone2005keepaway}, and the ``\textit{half field offense}'' \cite{kalyanakrishnan2006half} tasks).

While several successful case studies on multi-agent RL for soccer rely on high level actions, which incorporate domain knowledge (e.g. \cite{stone2005keepaway, kalyanakrishnan2006half,hausknecht2016deep}), recently, Liu et al. \cite{liu2018emergent} experimentally proved that using end-to-end multi-agent RL and decentralized population-based training (PBT) \cite{jaderberg2019human}, resulted on the emergence of collaborative behaviors in the 2v2 continuous soccer domain they proposed. Although promising results were obtained using PBT in this environment, the number of samples required to obtain proficient policies was extremely high (between 40B and 80B samples) \cite{liu2018emergent}.

In this work, we investigate the possibility of learning collaborative behaviors for playing soccer through multi-agent RL in a sample-efficient manner. We hypothesize that, although using end-to-end RL may allow the emergence of collaborative behaviors \cite{jaderberg2019human,liu2018emergent}, the incorporation of explicit curricula for learning, combined with competition, may allow to achieve this goal in a sample-efficient way.

Thus, we propose a multi-agent variant of the Twin Delayed Deep Deterministic Policy Gradient (TD3) algorithm \cite{fujimoto2018addressing} along with an explicit curriculum, and a competition-based training scheme, to address the 2v2 soccer problem introduced in \cite{liu2018emergent}. We divide the training process into three stages of increasing complexity: 1v0, 1v1, and finally 2v2. 

While the first stage (1v0) is modeled as a single-agent control task, and framed as an RL problem, the following two stages (1v1 and 2v2) are set in multi-agent environments, and the players/teams learn through competition against the expert agent/teams that are obtained from the corresponding previous stage, respectively (1v0 precedes 1v1, and 1v1 precedes 2v2). 

To speedup the learning process during the 1v1 and 2v2 stages, we use a basic form of ``\textit{experience sharing}'' (ES) \cite{tan1993multi} in which experiences that result from the interaction between the expert agents and the environment, are combined with those that result from the interaction between the agents being trained and the environment. Additionally, we show that incorporating a form of ``\textit{frame skipping}'' (FS) \cite{mnih2015human} increases the final performance of the trained soccer team. 

With the above, we experimentally show that complex skills required for playing 2v2 soccer proficiently, such as dribbling, feinting, intercepting the ball, and displaying a coordinated team play, may be learned from scratch in a competition-based setting. The obtained results also show that the proposed method reduces the number of interactions required for acquiring these skills by a factor of 1000x when compared to the PBT scheme proposed in \cite{liu2018emergent}.

%% file: related_work.tex
Learning to play soccer using RL has been a longstanding challenge. As a result, several successful case studies have been reported through the years, in both simulated and real-world environments. Some of these studies involve, for instance, the acquisition of skills required to perform sub-tasks of the full soccer problem, such as dribbling \cite{leottau2014ball}, scoring goals \cite{asada1994vision1, asada1994vision2}, and performing well in simplified settings of the game (e.g. in the ``\textit{keepaway}'' \cite{stone2005keepaway,stone2005reinforcement} and the ``\textit{half field offense}'' \cite{kalyanakrishnan2006half, hausknecht2016deep} tasks, and in matches with simpler rules and a reduced number of players \cite{wiering1999reinforcement, liu2018emergent, ocana2019cooperative}).

In this work, we focus on learning behaviors for playing soccer. While great progress has been made in the area, most of the research on the topic assumes the availability of high level actions (such as kicking and passing the ball). Furthermore, a great deal of expert knowledge is often introduced to get desirable results.

Recently, efforts to address problems set in multi-agent environments, in an end-to-end manner, have been reported. In \cite{jaderberg2019human}, end-to-end multi-agent PBT was used to train agents to play capture the flag. In \cite{liu2018emergent}, the approach proposed in \cite{jaderberg2019human} was adapted to learn proficient policies for playing soccer in a 2v2 setting. The results obtained by using a PBT scheme, both in \cite{jaderberg2019human} and \cite{liu2018emergent}, showed that the trained agents displayed a great degree of coordination, which spontaneously emerged.

A disadvantage of the method proposed in \cite{liu2018emergent}, is its extremely high computational cost, requiring at least 40B interactions to obtain proficient policies. Therefore, the use of a more sample-efficient multi-agent algorithm is highly desirable. In continuous-control, DDPG \cite{lillicrap2019continuous} and TD3 \cite{fujimoto2018addressing}, are commonly used, efficient RL algorithms. Multi-agent extensions of these have been proposed in \cite{lowe2017multi}, and \cite{ackermann2019reducing} respectively. These algorithms showed promising results in competitive-cooperative tasks, thus, making them suitable options for learning in complex, multi-agent environments.

Curriculum-learning, or the division of a task in a curriculum \cite{elman1993learning,bengio2009curriculum}, has been used in multi-agent RL settings. The advantage of this scheme is that resulting stages of the curriculum might be simpler tasks than the original one, and thus, this simplification may carry over to a simpler learning process. Curriculum-learning in the form of tournaments has been used previously in \cite{al2017continuous}. This form of competition-based learning allows the acquisition of skills that actually result in better competition performance.

%% file: methodology.tex
\subsection{Problem Formulation}

This work addresses the problem of playing soccer in a 2v2 setting, in which two teams of robots play soccer using the ``\textit{sudden death}'' format (the first team that scores wins the match). We model this problem as a Markov game defined by a set of states $\mathcal{S}$, $N$ agents, their respective observation and action sets, $\Omega^1, ..., \Omega^N$ and $\mathcal{A}^1, ..., \mathcal{A}^N$, their respective reward and observation functions, $\mathcal{R}^1, ..., \mathcal{R}^N$ and $\mathcal{O}^1, ..., \mathcal{O}^N$, a transition function $p(s_{t+1}|s_t, a^1_t,...,a^N_t)$, and an initial state distribution $p(s_1)$. At every time step $t$, each agent $i$ observes $o^i_t$, executes an action $a^i_t$ according to its policy $\pi^i$, and receives a scalar reward $r^i_t$. The environment then evolves to a new state $s_{t+1}$ according to the transition function. Each agent tries to maximize their respective expected discounted return $\mathbb{E}[\sum^T_{t=1} \gamma^{t-1} r^i_t]$. Compared to previous work, in which discrete action sets are often used (e.g. \cite{stone2005keepaway, kalyanakrishnan2006half, wiering1999reinforcement}), in this work both the state and action sets are continuous, and the task is naturally episodic, so $T$ is finite.

\subsection{Curriculum-Learning}
\label{subsec:curriculum-learning}

We divide the task of playing 2v2 soccer in three stages of increasing complexity: 1v0, 1v1, and finally 2v2. We use agents trained in stage $k$, as fixed opponents for the agents trained in stage $k+1$. 

In the first stage (1v0), a single agent learns to maneuver itself to score a goal. In this stage, the agent learns skills such as getting close to the ball, dribbling, and kicking the ball towards a goalpost. In the second stage (1v1), the agent learns to play against the policy trained in the previous stage (1v0), learning additionally to chase, intercept and feint. In the third and final stage (2v2), a team of two agents learns to play against a team of two independent agents trained in the second stage (1v1). As the opponents of the final stage cannot coordinate (their trained policies do not consider the presence of a teammate), the team being trained must learn some form of coordination to exploit the other team's weakness.

\subsection{Experience Sharing}

Under the curriculum described above, agents are trained on stages of increasing difficulty. Transferring knowledge across stages can be particularly useful in this scenario. This idea may be hard to apply in a number of tasks, especially in those in which every stage has a different observation space, so policies trained in a given stage cannot be retrained directly on the next stage. Thus, another approach for skill transfer is required. 

In this work, we propose using transitions experienced by fixed opponent players in the current stage, to speed up the learning process of the agent being trained. Given that fixed opponent players were trained in the previous stage, knowledge is transferred across stages. This may be interpreted as the simplest form of experience sharing (ES). The effect of ES is two-fold, on one hand, the agent quickly learns what actions offer a better reward than those obtained in early stages, avoiding the need for heavy exploration. On the other hand, ES eases the agent the acquisition of baseline behaviors that are required to at least match the opponent's performance.

\subsection{Actions}

The $i$-th agent's actions correspond to 3-dimensional vectors, $a^i_t \in [-1, 1]^3$. Each component of $a^i_t$ represents the linear acceleration, the torque on the vertical axis that allows rotation, and a downwards force that can be used to make the agent jump, respectively. For the sake of simplicity, the third component of each action (the downwards force) is fixed to zero, thus, forcing the agents to stay on the ground. This simplification is also in line with the fact that robots in soccer leagues currently are unable to jump.

\subsection{Observations}

The $i$-th agent's observations, $o^i_t$, consist mainly of 2-dimensional position, velocity and acceleration vectors. These observations can be divided into two groups: the first group contains proprioceptive measurements, and information related to the position of key points in the field with respect to its local frame, while the second group contains information about its teammates and opponents in the field. The components that conform the agents' observations are listed in Table \ref{tab:obs}. 

All position vectors are transformed to their polar form, i.e. to a distance and an angle. The distance is normalized by the maximum measurable distance (the field diagonal length), and the angle is normalized by $2\pi$. On the other hand, velocity and acceleration vectors are also transformed to a modified polar form: the angle is obtained and normalized as described above, while a modified scaled magnitude, $|\overline{\rho}|$, is computed as $\sqrt{(\tanh^2(c_x) + \tanh^2(c_y))/2}$, where $c_x$ and $c_y$ are the $x$ and $y$ components for the velocity or acceleration, as appropriate.

Additional information, such as whether the ball is or ever has been at a kick-able distance, and the projection of the ball's velocity on the agent to opponent goalpost vector, are also components of the observations. The former is a boolean value, and thus is casted to either 0 or 1, while the latter, which is a signed scalar, is normalized with a sigmoid function.

\begin{table}[]
    \caption{Components of the agents' observations}
    \label{tab:obs}
    \centering
    \begin{adjustbox}{max width=\linewidth}
    \begin{tabular}{@{}clc@{}}
        \toprule
        \textbf{Component} &\textbf{Description} & \textbf{Dimensions}  \\ \midrule
        $o^i_{\text{vel}}$ & Agent's velocity                                            & 2 \\ \midrule
        $o^i_{\text{acc}}$ & Agent's acceleration                                        & 2 \\ \midrule
        $o^i_{\text{b}_\text{pos}}$ & Local ball position                                         & 2 \\ \midrule
        $o^i_{\text{b}_\text{vel}}$ & Local ball velocity                                         & 2 \\ \midrule
        $o^i_{\text{op}_\text{gp}}$ & Local opponent goalpost position                            & 2 \\ \midrule
        $o^i_{\text{tm}_\text{gp}}$ & Local team goalpost position                                & 2 \\ \midrule
        \multirow{2}{*}{$o^i_{\text{b}_\text{pos}} - o^i_{\text{op}_\text{gp}}$}  & Difference between local ball position, & \multirow{2}{*}{2} \\ 
        & and local opponent goalpost position & \\\midrule
        \multirow{2}{*}{$o^i_{\text{b}_\text{pos}} - o^i_{\text{tm}_\text{gp}}$} & Difference between local ball position, & \multirow{2}{*}{2}\\
         & and local teammate goalpost position& \\\midrule
        \multirow{2}{*}{$\left(\text{proj}(o^i_{\text{b}_\text{vel}},o^i_{\text{op}_\text{gp}}), o^i_{\text{kick}}\right)$} & Projected ball velocity, and boolean  & \multirow{2}{*}{2} \\
         & for ``ball is or has been kick-able''                                 &  \\ \midrule
        \multirow{4}{*}{$\left(o^i_{\text{j}_\text{pos}}, o^i_{\text{j}_\text{vel}}, o^i_{\text{j}_\text{pos}}-o^i_{\text{b}_\text{pos}}\right)$} & $j$-th agent local position,                                   & \multirow{4}{*}{6} \\
         & $j$-th agent local velocity, and                                 &  \\
         & difference between $j$-th agent local &  \\
          & position and the ball's agent local position & \\
        \bottomrule
    \end{tabular}
    \end{adjustbox}
\end{table}

\subsection{Reward Functions}

To guide the agent's learning process, a hand-crafted dense reward function is designed. The effect of using this reward function is compared against using sparse rewards. Both variants are described below.

\subsubsection{Dense Reward Function}

This reward function specifically enforces sub-tasks that might be essential for learning to play soccer: it is designed to guide the agent to first get close to the ball, and once close enough, to kick or dribble the ball towards the opponent's goalpost, while avoiding to get it closer to the agent's own goalpost. 

To properly describe this function, the following values are defined:

\begin{itemize}
    \item $\alpha$: Max. number of steps in an episode, divided by 10.
    \item $\beta$: $\alpha/10$.
    \item $\lambda$: Normalized distance threshold (in our experiments, this distance is set to 0.03).
    \item $d^i_t$: Normalized distance of the $i$-th agent to the ball at time step $t$.
    \item $D^l_t$: Normalized distance of the ball to the center of the goalpost where team $l\in \{0, 1\}$ should score.
    \item $b^i_t$: Boolean, \texttt{true} if $d^i_{t^\ast} \leq \lambda$ for some $t^\ast < t$, \texttt{false} otherwise. Represents whether the ball has been at a kick-able distance before.
    \item $k^i_t$: Boolean, \texttt{true} if $b^i_t$ is \texttt{false} and $d^i_t \leq \lambda$, \texttt{false} otherwise. Represents whether the ball is at a kick-able distance for the first time.
\end{itemize}

Given the values defined above, the reward for player $i$ belonging to team $l \in \{0, 1\}$, at time step $t$, can be obtained according to Eq. \eqref{eq:rew}, where the terms $r^{\text{\xmark}}_t$ and $r^{\text{\cmark}}_t$ are defined by Eqs. \eqref{eq:rew-nogoal}\footnote{$\Delta \xi_t \coloneqq (\xi_t - \xi_{t-1})$, where $\xi_{t^\ast}$ corresponds to a measured distance at time step $t^\ast$.} and \eqref{eq:rew-goal}, respectively.

\begin{equation}
    \label{eq:rew}
    r_t = \left\{\begin{array}{ll}
         r^{\text{\xmark}}_t &  \text{if a goal has not been scored},\\
         r^{\text{\cmark}}_t &  \text{otherwise}.
    \end{array}
    \right.
\end{equation}

\begin{equation}
    \label{eq:rew-nogoal}
    r^{\text{\xmark}}_t= \left\{\begin{array}{ll}
         \beta - 0.1 & \text{if } k^i_t, \\
         1.2 \cdot (\Delta D^{1-l}_t - \Delta D^l_t) - \Delta d^i_t - 0.1 & \text{if } b^i_t, \\
         -\Delta d^i_t - 0.1 & \text{otherwise.} \\
    \end{array}\right.
\end{equation}

\begin{equation}
    \label{eq:rew-goal}
    r^{\text{\cmark}}_t = \left\{\begin{array}{ll}
         +\alpha & \text{if goal scored in team $1-l$'s goalpost,} \\
         -\alpha & \text{if goal scored in team $l$'s goalpost.}
    \end{array}
    \right. 
\end{equation}

While a goal has not been scored, $r_t$ equals the value of $r^{\text{\xmark}}_t$. In this scenario, three conditions are considered. When the ball is at a kick-able distance for the first time ($k^i_t$ equals \texttt{true}), the agent receives a significant reward. For the following time steps ($b^i_t$ equals \texttt{true}) the function rewards kicks or dribbles if they decrease the distance between the ball and the opponent's goalpost, whilst actions that get the ball close to the team's own goalpost, or move the agent far from the ball, are penalized. If both of the previous conditions have not been met ($\lnot (b^i_t \vee k^i_t)$ equals \texttt{true}), then the agent is rewarded for getting close to the ball.

When a goal is scored, $r_t$ equals the value of $r^{\text{\cmark}}_t$, so the scoring team is given a large reward, whereas the defeated team receives a large punishment.

\subsubsection{Sparse Reward Function}
In this case, the reward is given by Eq. \eqref{eq:rew-sparse}, so the agent is only rewarded or punished depending on the final outcome of a match.

\begin{equation}
    \label{eq:rew-sparse}
    r_t = \left\{\begin{array}{ll}
         +1 & \text{if goal scored in team $1-l$'s goalpost,} \\
         -1 & \text{if goal scored in team $l$'s goalpost,} \\
         0 & \text{otherwise.}
    \end{array}
    \right.
\end{equation}

\subsection{Learning Algorithm}

\subsubsection{Multi-Agent TD3}
Twin-Delayed Deep Deterministic Policy Gradient (TD3) was proposed by Fujimoto et al. in \cite{fujimoto2018addressing}, built upon the Deep Deterministic Policy Gradient (DDPG) algorithm \cite{lillicrap2019continuous}.

TD3 incorporates various improvements that allow a faster convergence, while reducing the degree of value function overestimation \cite{fujimoto2018addressing}. To adapt this method to a multi-agent setting, the simplest approach is followed: we use separate actor and critic networks, and independent replay buffers for every agent. 

The proposed method is shown in Algorithm \ref{algo:matd3+se}, the steps associated with ES are displayed in blue. These steps involve sampling transitions experienced by the fixed expert opponent players, and using them along with the agent's own expercienced transitions for training.

\subsubsection{Actor and Critic Networks}

\begin{figure}
    \centering
    \includegraphics[width=0.48\textwidth]{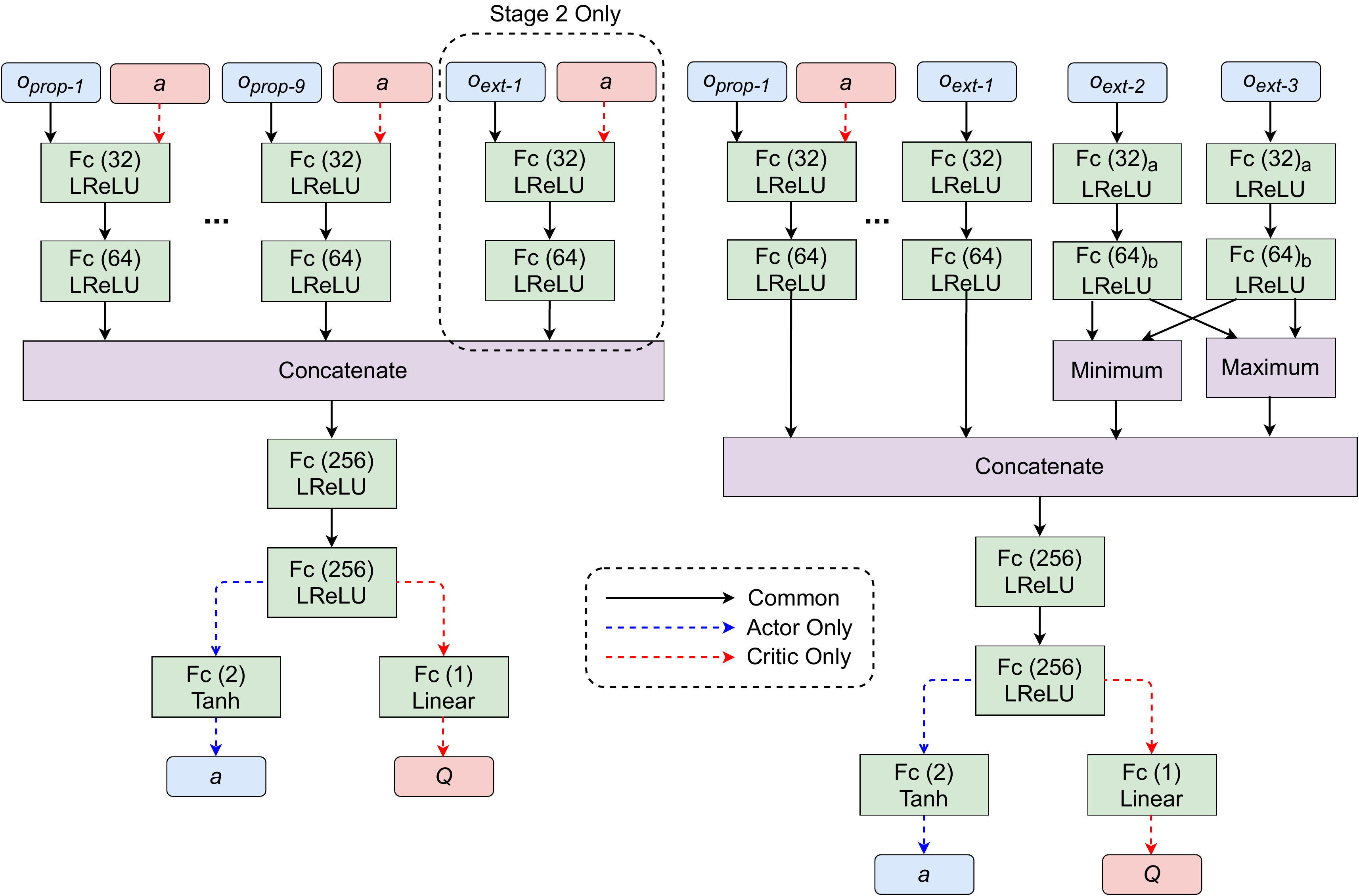}
   
    \caption{Architectures for both the actor and critic networks. The architecture for stages 1 and 2 is shown on the left, while the architecture for stage 3 is shown on the right. In the latter network, layers with the same subscripts ($a$ and $b$) share weights. Red input cells represent the agent's action, blue input cells represent the agent's observations, green cells correspond to intermediate layers, blue output cells correspond to the action selected by the actor, and red output cells represent the estimates provided by the critic.}
    \label{fig:archs}
\end{figure}

The architectures for the actor and critic networks are shown in Fig. \ref{fig:archs}. Each proprioceptive observation's component displayed in Table \ref{tab:obs} (rows one to nine) is denoted as $o^i_\text{prop-$\delta$}$, $\delta=1,...,9$, while the exteroceptive component (row 10) is denoted as $o^i_\text{ext-$j$}$, $1\leq j \leq N-1$, where $N$ is the total number of agents.

These architectures are designed so they allow an equal importance of every component of the observations, while assigning a prominent relevance to the actions for the Q function estimates. This design decision follows some insights provided in \cite{heess2017emergence}.

Different network architectures are used for the stages considered in the defined curriculum (see Section \ref{subsec:curriculum-learning}). For stages 1 and 2, simpler architectures are used (see Fig. \ref{fig:archs} left), while for stage 3, modifications are introduced to make the networks invariant to the order in which the opponent observations are fed. Features associated to both the opponent player observations are obtained with shared weights, and then the element-wise minimum and maximum are concatenated with the rest of the intermediate representations, in a similar fashion to what is done, for instance, in \cite{liu2018emergent} or in \cite{huttenrauch2019deep}.



\begingroup
\removelatexerror
\begin{algorithm}[H]
\caption{Proposed Multi-Agent TD3 with ES}
\label{algo:matd3+se}
$N_{\text{train}}$: Number of players to be trained \\ $N_{\text{total}}$: Total number of players in a match \\
$M$: Batch size\\
\textcolor{blue}{$M'$: Sample size from each buffer, $\Big\lfloor \frac{M}{N_{\text{total}} - N_{\text{train}} + 1}\Big\rfloor$}\\
\For{$i=1$ \textbf{to} $N_{\text{train}}$}{
    Initialize critics $\theta_{1,i}$, $\theta_{2,i}$, actor $\phi_i$, target networks $\theta'_{1,i} \leftarrow \theta_{1,i}$, $\theta'_{2,i} \leftarrow \theta_{2,i}$, $\phi'_i \leftarrow \phi_i$, and  replay buffer $\mathcal{B}_i$
}
\For{$t=1$ {\bfseries to} $T_\text{train}$}{
    \For{$i=1$ {\bfseries to} $N_{\text{train}}$}{
        \eIf{$t \leq T_\text{warmup}$}{
            Select action $a_i \sim \text{Uniform}(a_{\text{low}}, a_{\text{high}})$
        }{
            Select action $a_i \sim \pi_{\phi_i}(s_i) + \epsilon$, $\epsilon \sim \mathcal{N}(0, \sigma)$
        }
  }
  \For{$i=N_{\text{train}} + 1$ {\bfseries to} $N_{\text{total}}$}{
        Select action $a_i \sim \pi_{\phi_i}(s_i)$
  }
  Apply actions, observe rewards and new states. \\
  \For{$i=1$ {\bfseries to} $N_{total}$}{
        Store transition tuple $(s_i, a_i, r_i, s'_i)$ in $\mathcal{B}_i$
  }
  \If{($t$ mod $u$ equals $0$) and $t \geq T_{\text{after}}$}{
        \For{$i=1$ {\bfseries to} $N_{\text{train}}$}{
            Sample mini-batch of $M'$ transitions $(s, a, r, s')$ from $\mathcal{B}_{i}$ \\
            \textcolor{blue}{
            \For{$j=N_{train} + 1$ {\bfseries to} $N_{total}$}{
                Sample mini-batch of $M'$ transitions $(s, a, r, s')$ from $\mathcal{B}_{j}$
            }}
            $\tilde{a} \leftarrow \pi_{\phi'_{i}}(s') + \epsilon, \epsilon \sim \text{clip}(\mathcal{N}(0, \tilde{\sigma}), -c, c)$ \\
            $y \leftarrow r + \gamma \min_{n=1,2} Q_{\theta'_{n,i}}(s', \tilde a)$\\
            \For{$z=1$ {\bfseries to} $u$}{
                Update critics $(n =1, 2)$:
                $\theta_{n,i}\leftarrow \text{argmin}_{\theta_{n,i}} M^{-1} \sum (y - Q_{\theta_{n,i}}(s,a))^2$ \\
                \If{$z$ mod $d$}{
                    Update $\phi_i$ by the deterministic policy gradient ($n = 1, 2$): $\nabla_{\phi_i} J(\phi_i) = M^{-1} \sum \nabla_{a} Q_{\theta_{n,i}}(s, a) |_{a=\pi_{\phi_i}(s)}\cdot$ $\nabla_{\phi_i} \pi_{\phi_i}(s)$ \\
                    Update target networks ($n = 1, 2$): 
                    $\theta'_{n, i} \leftarrow \tau \theta_{n,i} + (1 - \tau) \theta'_{n,i}$
                    $\phi'_i \leftarrow \tau \phi_i + (1 - \tau) \phi'_i$
                }
            }
        }
    }
}
\end{algorithm}
\endgroup

\subsection{Training Procedure}

The training procedure is incremental and considers three stages, as indicated in Section \ref{subsec:curriculum-learning}. In stage 1 (1v0), the agent learns how to approach the ball, and how to score goals. In stage 2 (1v1), it learns how to play against an opponent. Finally, in stage 3 (2v2), two agents learn how to play against an opposing team.

\subsubsection{Stage 1 (1v0)}

This stage is akin to learning how to play soccer by oneself, i.e. the setting consists of a single agent, a ball, and a goalpost. The objective is to score a goal before reaching a certain time limit. Given that this task may be framed as a single-agent RL problem, using vanilla TD3 as a learning algorithm is enough in this case. 

\subsubsection{Stage 2 (1v1)}
\label{subsec:stage2}

In the previous stage (1v0), the resulting policy enables an agent to score a goal in an empty field. The aim of this stage is to endow an agent with the skills required to defeat agents trained in the 1v0 setting.

\subsubsection{Stage 3 (2v2)}

The aim of this stage is to train a team of two agents, each of them capable of observing their teammate, and the two opponents. The opponent team consists of two independent agents trained in stage 2 (1v1). It is important to note that this opposing team is incapable of coordinating its actions, as policies trained in stage 2 do not consider the presence of a teammate. 

This setting forces the trained agents to develop the necessary skills to defeat their opponents, given the competitive nature of this stage. Ideally, the team's agents must learn to use their teammate's and opponent's information to their advantage. 

Given that the policies trained in stage 2 (1v1) consider just one opponent, a scheme must be designed to decide which agent will observe each player of the trained team. Taking the simplest approach, i.e. every agent trained in stage 2 observes a fixed single opponent throughout the match, is sufficient to fulfill the aim of this stage.

\begin{figure*}[]
    \centering
    \includegraphics[width=0.55\linewidth]{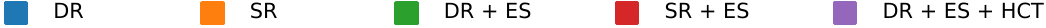}
    \\
    \subfloat[Stage 1 (1v0) \label{fig:sr-1v0}]{%
    \includegraphics[width=0.32\textwidth]{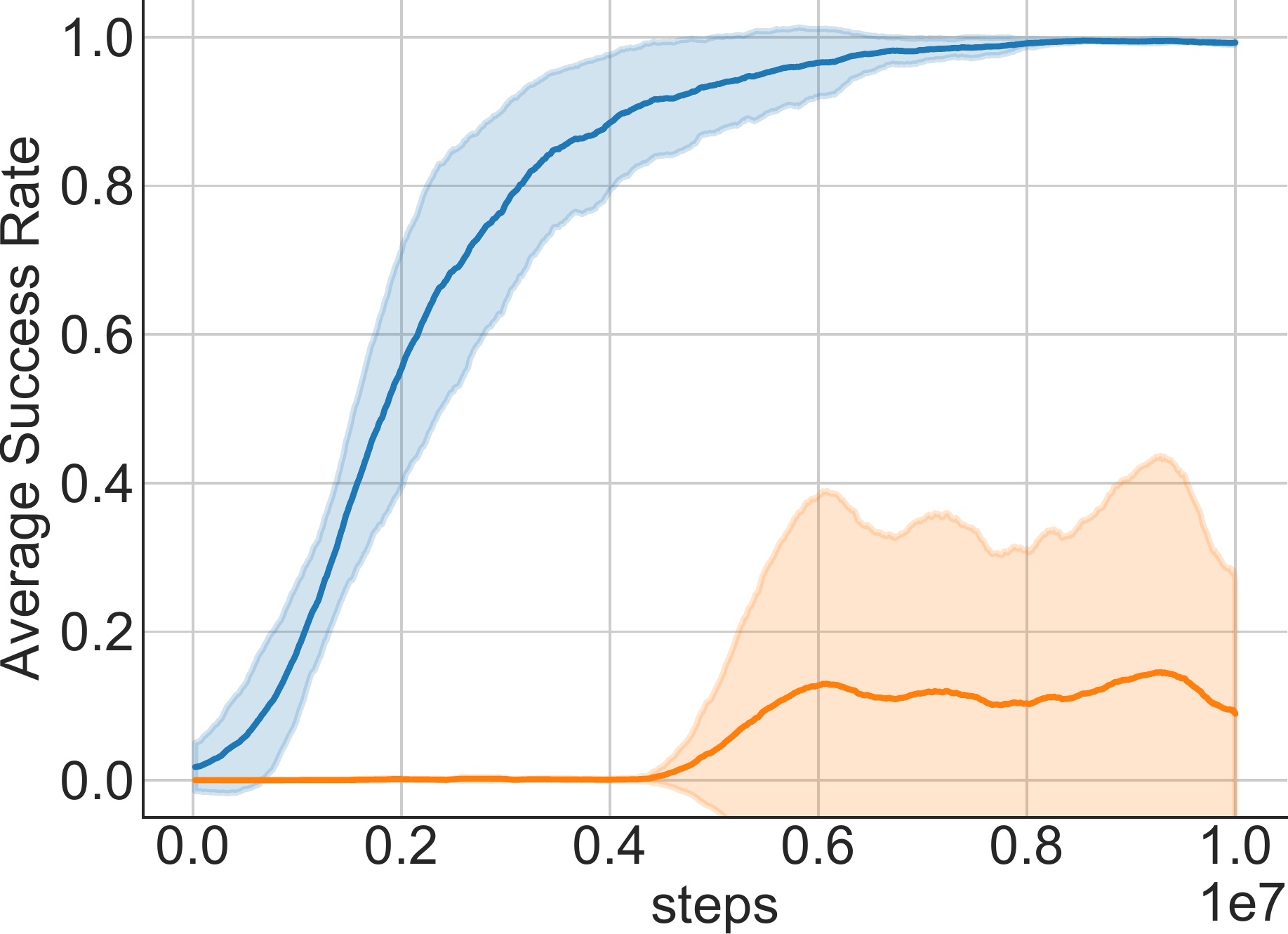}}\hfill
    \subfloat[Stage 2 (1v1) \label{fig:sr-1v1}]{%
    \includegraphics[width=0.32\textwidth]{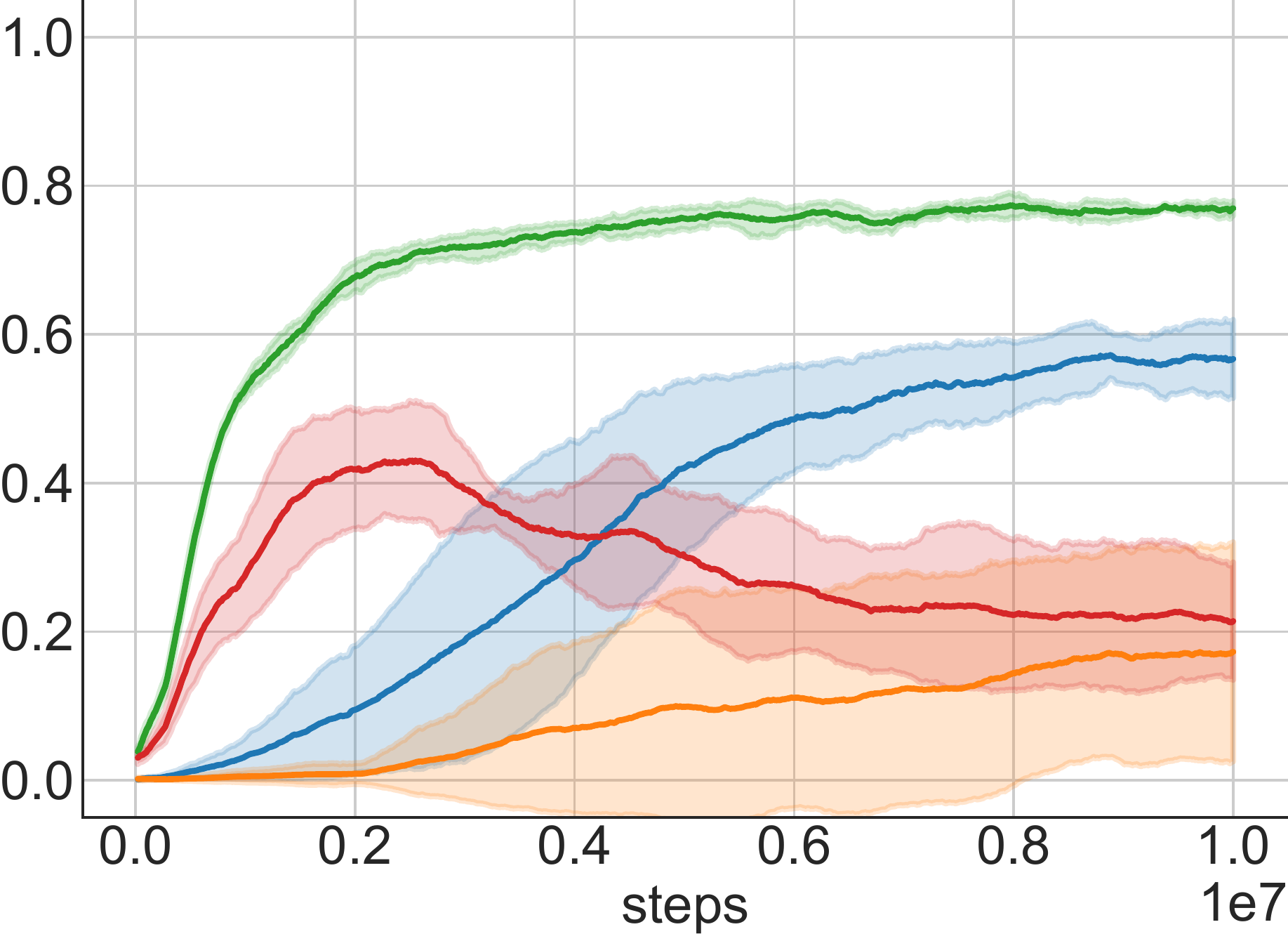}}\hfill
    \subfloat[Stage 3 (2v2) \label{fig:sr-2v2}]{%
    \includegraphics[width=0.32\textwidth]{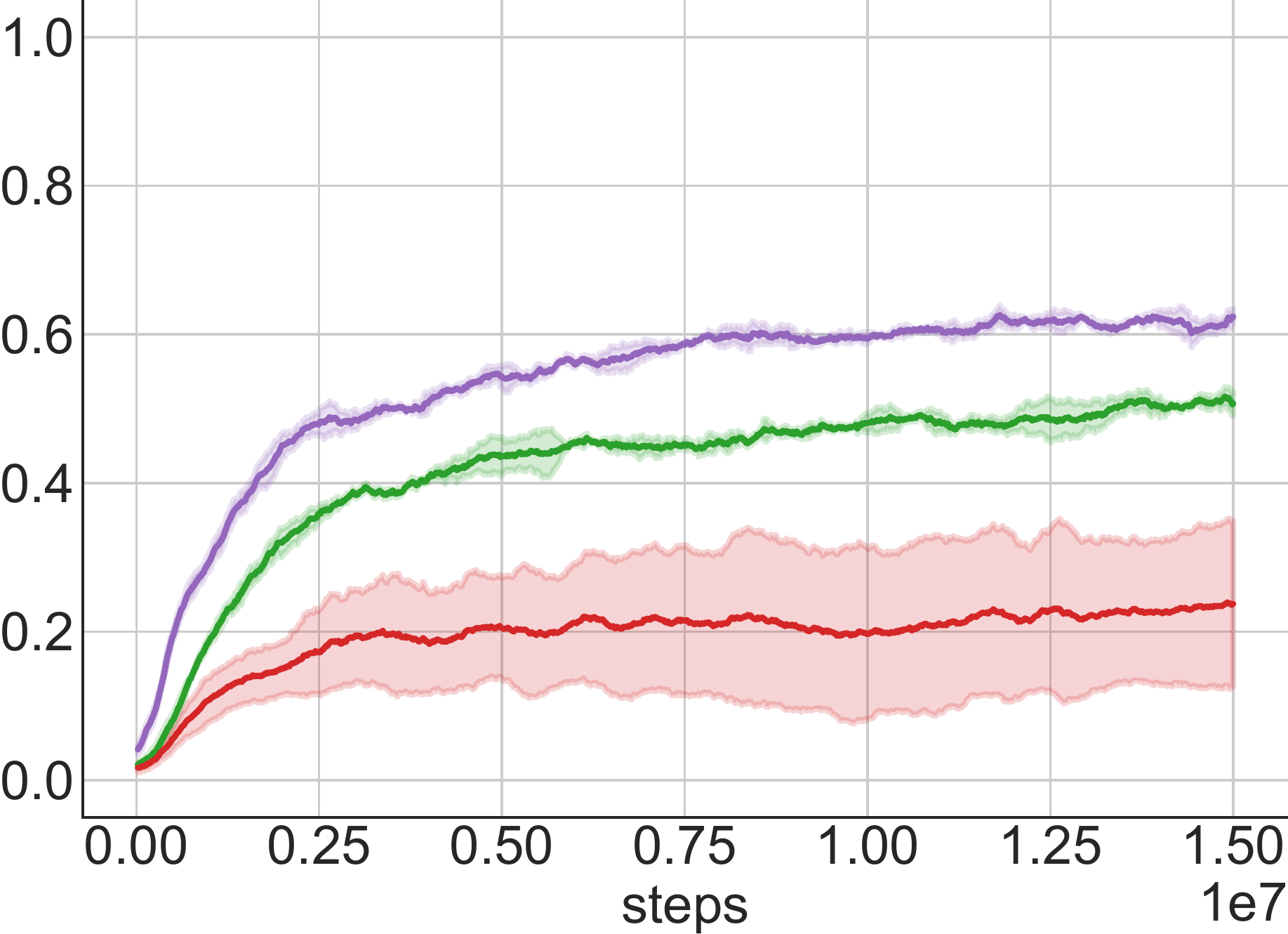}}
    \caption{Evolution of trained agent/team's success rate,  averaged over 5 random seeds (3 in the case of stage 3). Curves are smoothed using a window of size 50. DR: Dense Reward, SR: Sparse Reward, ES: Experience Sharing, HCT: Higher Control Time step.}
    \label{fig:sr}
\end{figure*}

\subsection{Agent Selection}
\label{subsec:meth_agent_selection}

An important decision to be made is which agent trained in stage $k$ should be selected as the fixed opponent for stage $k+1$. To measure the performance of every agent, we use Nash Averaging \cite{balduzzi2018re}, given its property of invariancy to redundant agents, allowing unbiased comparisons with respect to the conventional ELO rating \cite{elo1978rating}. Nash Averaging is used to evaluate agents by computing the average payoff to be obtained by a meta-player when choosing a certain agent, when the opponent meta-player follows an optimum Nash correlated equilibrium strategy. The same approach was used in \cite{liu2018emergent} to evaluate performance.

To select which agents are used as fixed opponents in stages $k+1$ ($k=1, 2$), these are first filtered according to their performance on the task they were trained in (stage $k$), and then evaluated in stage $k + 1$. The pool of agents considered consists of all agents saved every $10,000$ time steps, during the last 20\% of the training process.

With the above, we define the following procedures for selecting the agents of stage $k$:

\begin{itemize}
    \item \textit{Stage 1 (1v0)}: The set of all agents with 100\% success rate on the task of scoring a goal within 30 seconds defines the initial pool of agents. Two metrics, the average episode length and the average \textit{vel-to-ball} (agent's velocity projected on the agent to ball vector) are recorded. 

    An agent $i$ is then considered to be pareto-dominant over an agent $j$, if it required, on average, a lesser number of steps to solve the task of scoring, and did so with a higher average \textit{vel-to-ball} metric. 

    Then, pareto-dominant individuals play soccer against each other in the 1v1 format. The resulting expected goal differences among agents are then used to define a payoff matrix and calculate the Nash rating of each agent. Finally, the agent with the highest Nash rating is selected as the fixed opponent for stage 2.
    
    \item \textit{Stage 2 (1v1)}: Agents with the top 95\% performance (success rate) on the task of playing soccer against the agent selected in stage 1, are initially selected. As in the previous stage, the average episode length and the average \textit{vel-to-ball} metrics are recorded. 
    
    Then, pareto-dominant individuals with respect to the two recorded metrics, form all possible two-player teams, which then compete against each other in the 2v2 format. 
    
    The same procedure for obtaining the Nash rating through the expected goal differences among resulting teams, is repeated for this stage. Finally, the team with the highest Nash rating is selected as a fixed opponent for stage 3.
\end{itemize}

%% file: results.tex
\begin{figure}
    \centering
    \subfloat[Stage 1 (1v0), $\text{DR}$ scheme. \label{fig:pareto-1v0}]{
    \includegraphics[width=0.498\linewidth]{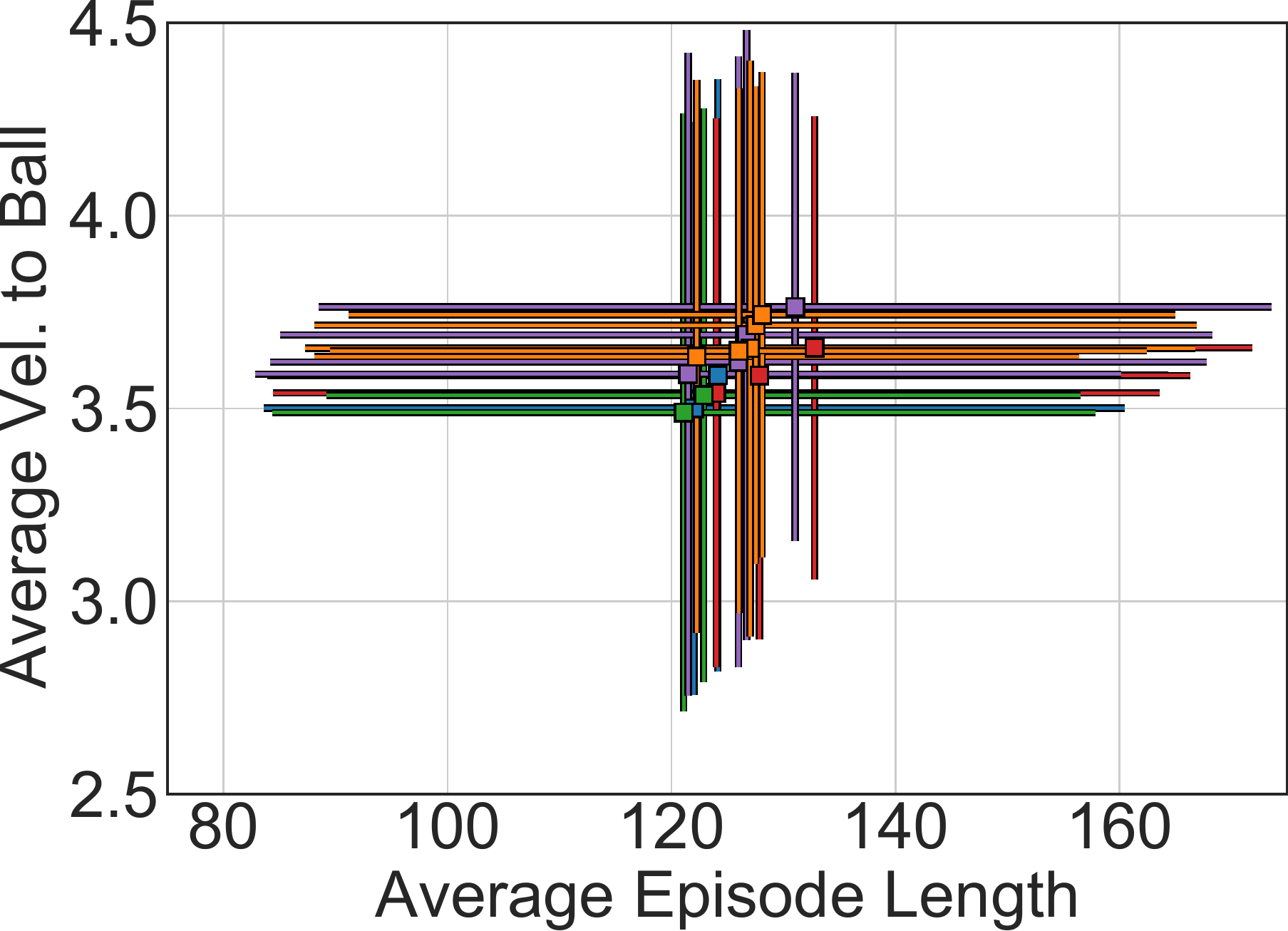}}\hfill
    \subfloat[Stage 2 (1v1), $\text{DR} + \text{ES}$ scheme. \label{fig:pareto-1v1}]{%
    \includegraphics[width=0.485\linewidth]{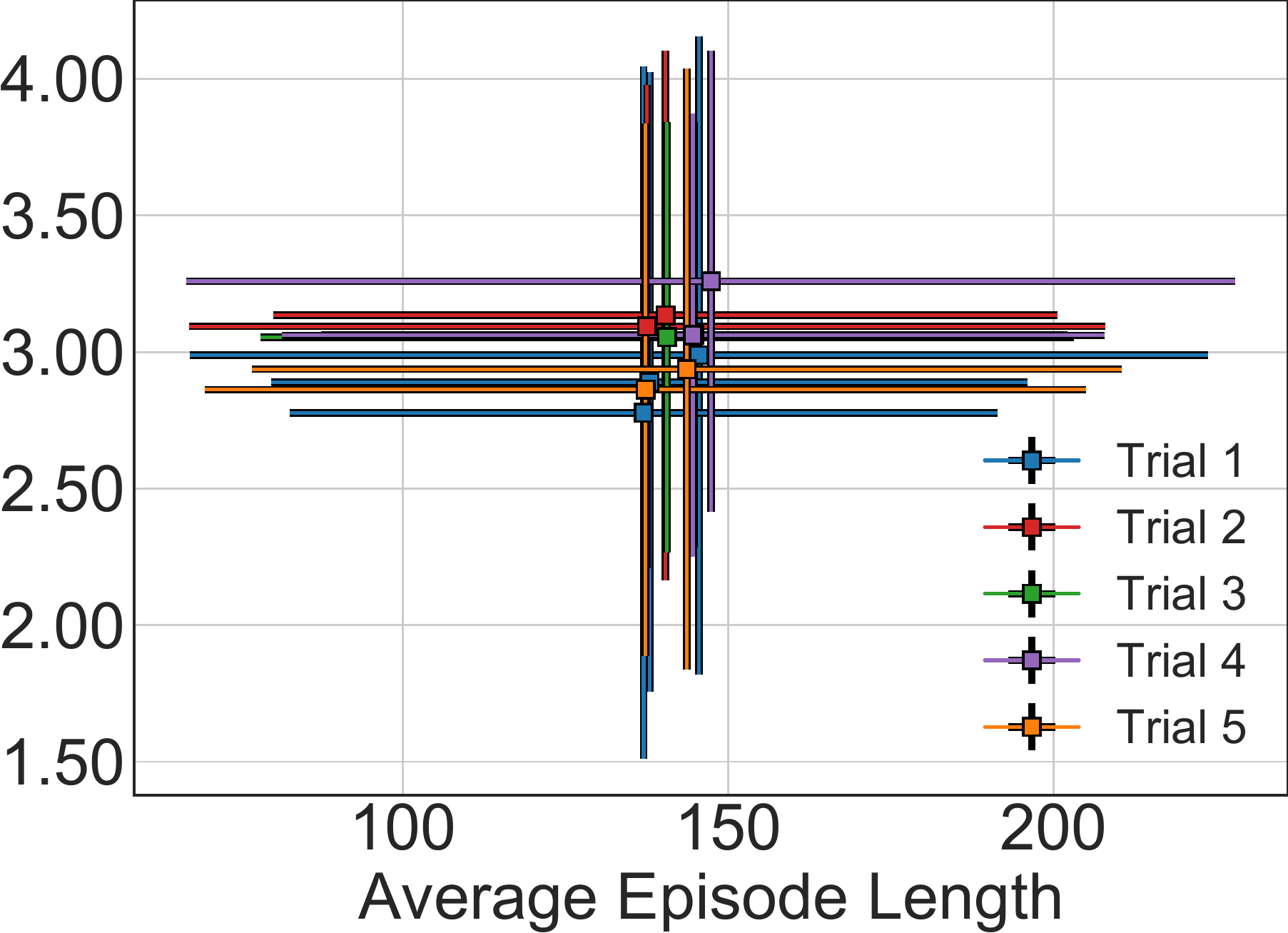}}\hfill
    \caption{Average and standard deviation of the \textit{vel-to-ball} and episode length metrics of dominant individuals obtained over 5 random seeds for stages 1 and 2. Each color depicts a different random seed. These individuals then compete against each other.}
    \label{fig:pareto}
\end{figure}

\begin{figure*}
    \centering
    \subfloat[Stage 1 (1v0). \label{fig:payoff-1v0}]{%
    \includegraphics[width=0.32\textwidth]{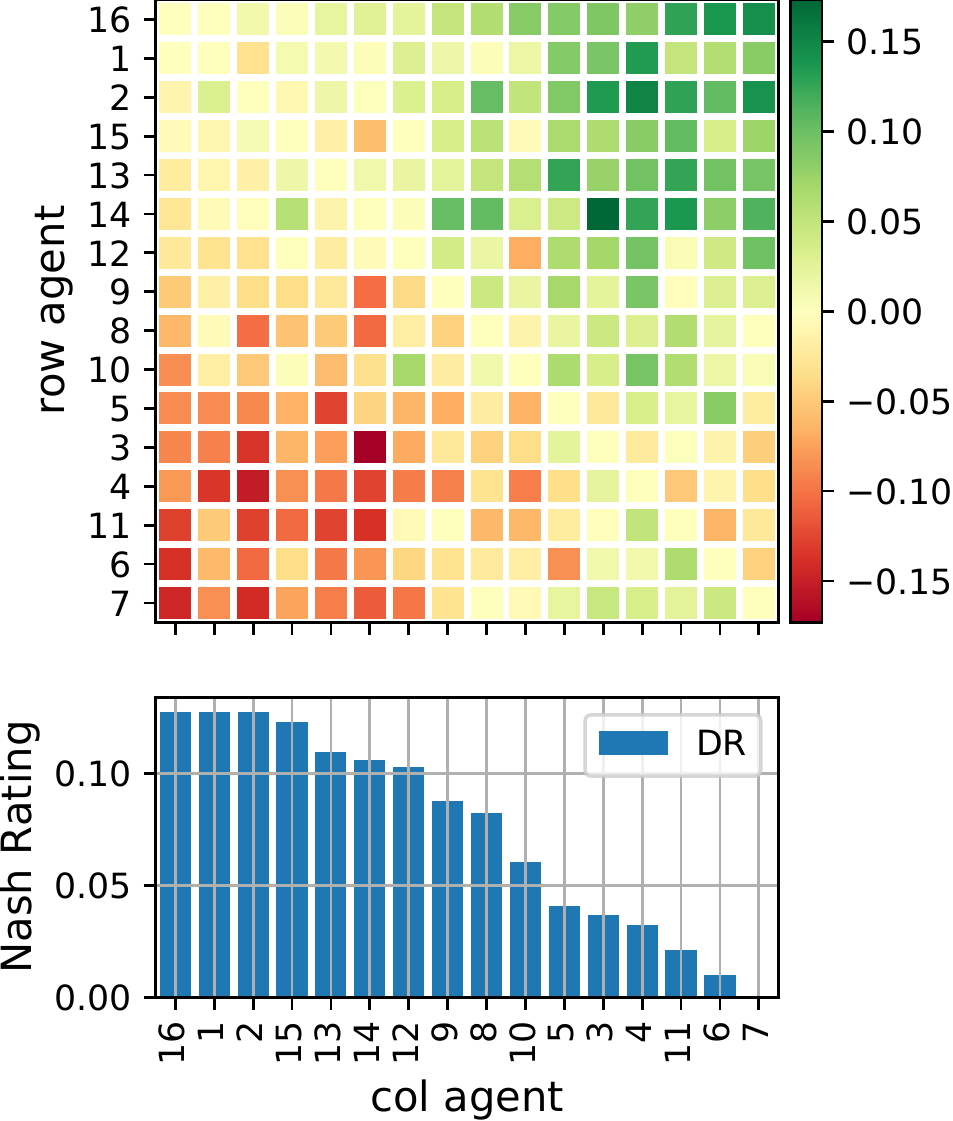}}\hfill
    \subfloat[Stage 2 (1v1). \label{fig:payoff-1v1}]{%
    \includegraphics[width=0.32\textwidth]{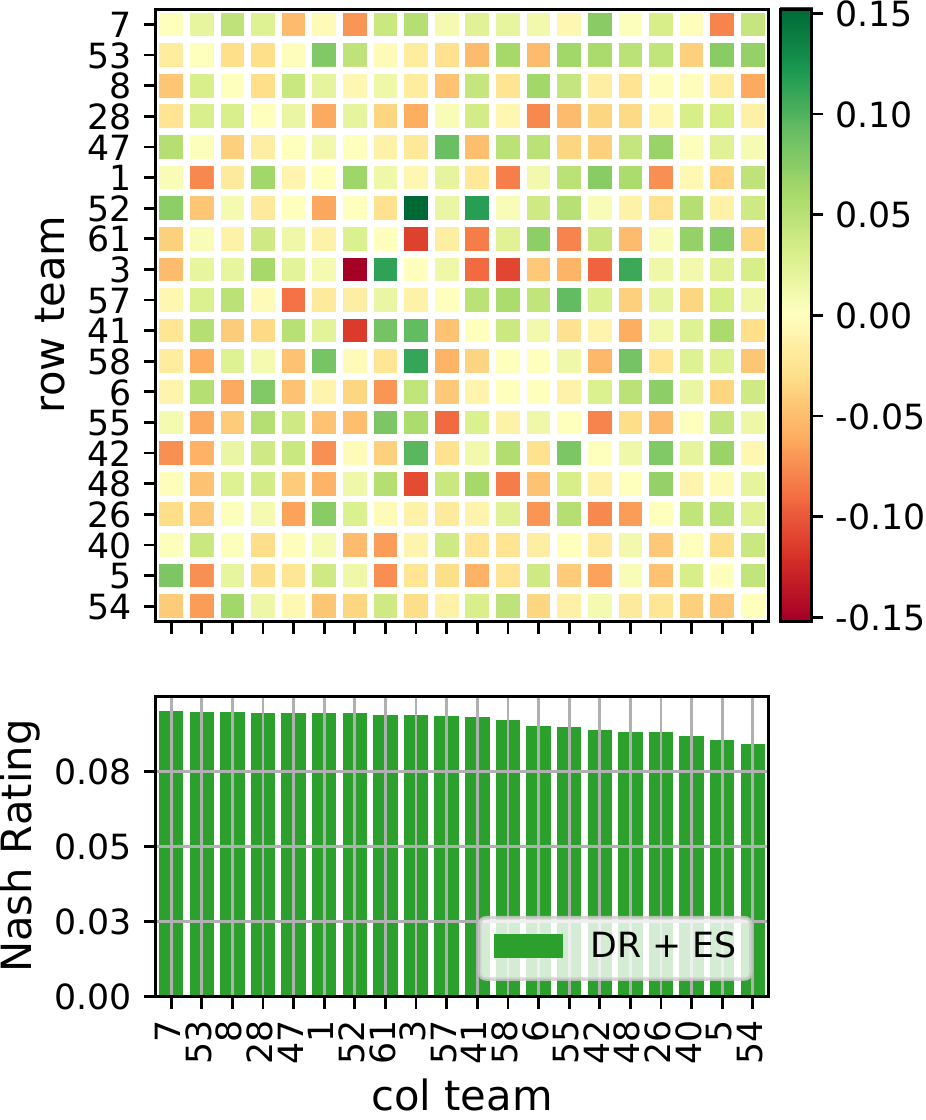}}\hfill
    \subfloat[Stage 3 (2v2). \label{fig:payoff-2v2}]{%
    \includegraphics[width=0.32\textwidth]{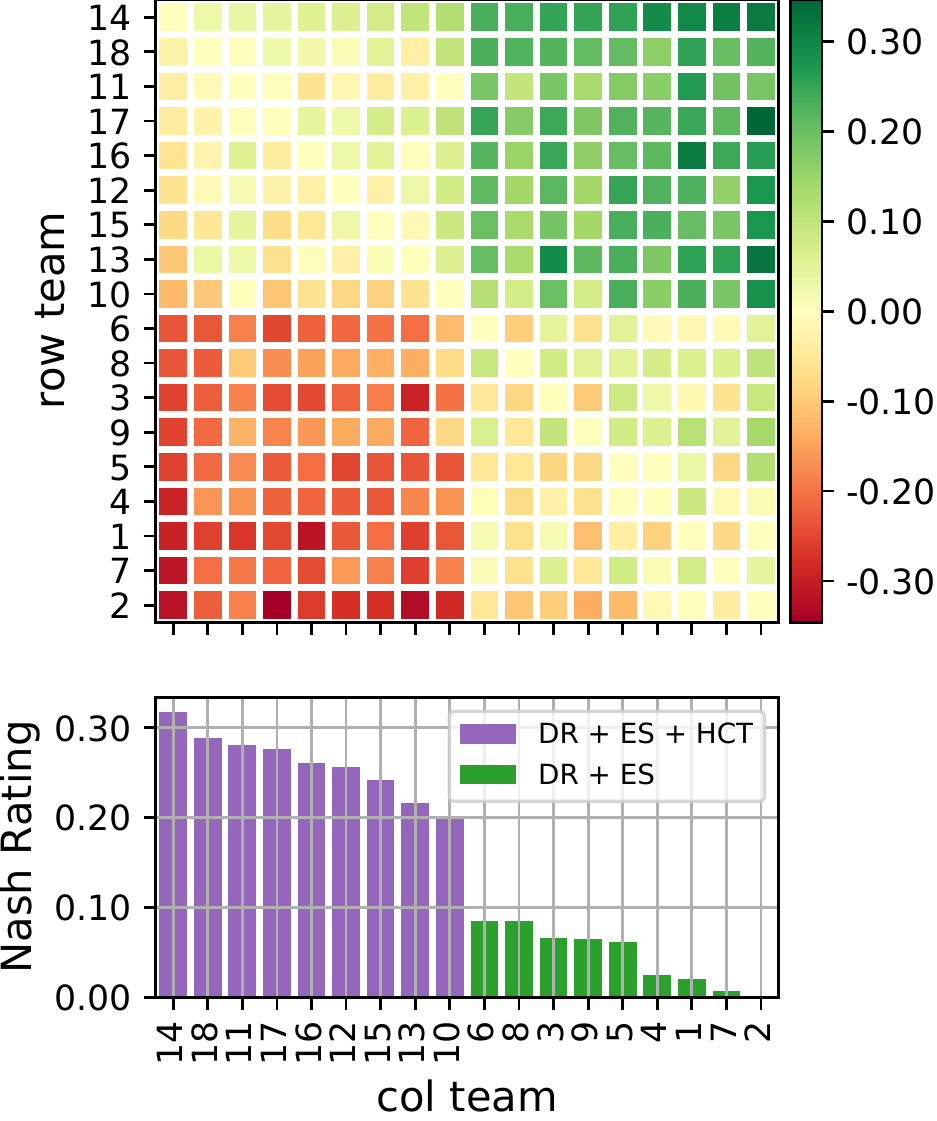}}
    \caption{Expected goal differences among best agents/teams in all training stages along with their Nash Rating. ($i$, $j$) (row, col) represents the expected goal-difference in favor of agent/team $i$ over $j$.}
    \label{fig:payoff}
\end{figure*}

Evolution of the success rate on each stage is shown in Figure \ref{fig:sr}; results for stages 1, 2, and 3 are shown in Figures \ref{fig:sr-1v0}, \ref{fig:sr-1v1} and \ref{fig:sr-2v2}, respectively. It can be observed that success rates obtained when using a dense reward (DR) are significantly higher than those obtained when using the sparse reward (SR) we consider. This confirms that the proposed DR eases the acquisition of skills required to play soccer.

\subsection{Experience Sharing}

Experience sharing (ES) was used while training in stages 2 and 3. This was done by using transition tuples $(s, a, r, s')$ experienced by agents trained in stages 1 and 2, when they were used as fixed opponents in stages 2 and 3, respectively.

As shown in Figure \ref{fig:sr-1v1}, ES increases performance and reduces variance. It can be seen that incorporating ES when using DR increases the success rate of the trained agent by 20\% in the task of 1v1 soccer.

\subsection{Effect of Control Time Step}

 \begin{figure*}
    \centering
    \includegraphics[width=0.75\textwidth]{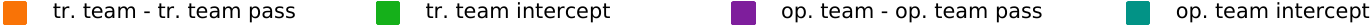}
    \\
    \subfloat[Evolution of performance metrics for team trained in Stage 3 (2v2) under the $\text{DR} + \text{ES}$ scheme. \label{fig:metrics-}]{%
    \includegraphics[width=0.24\textwidth]{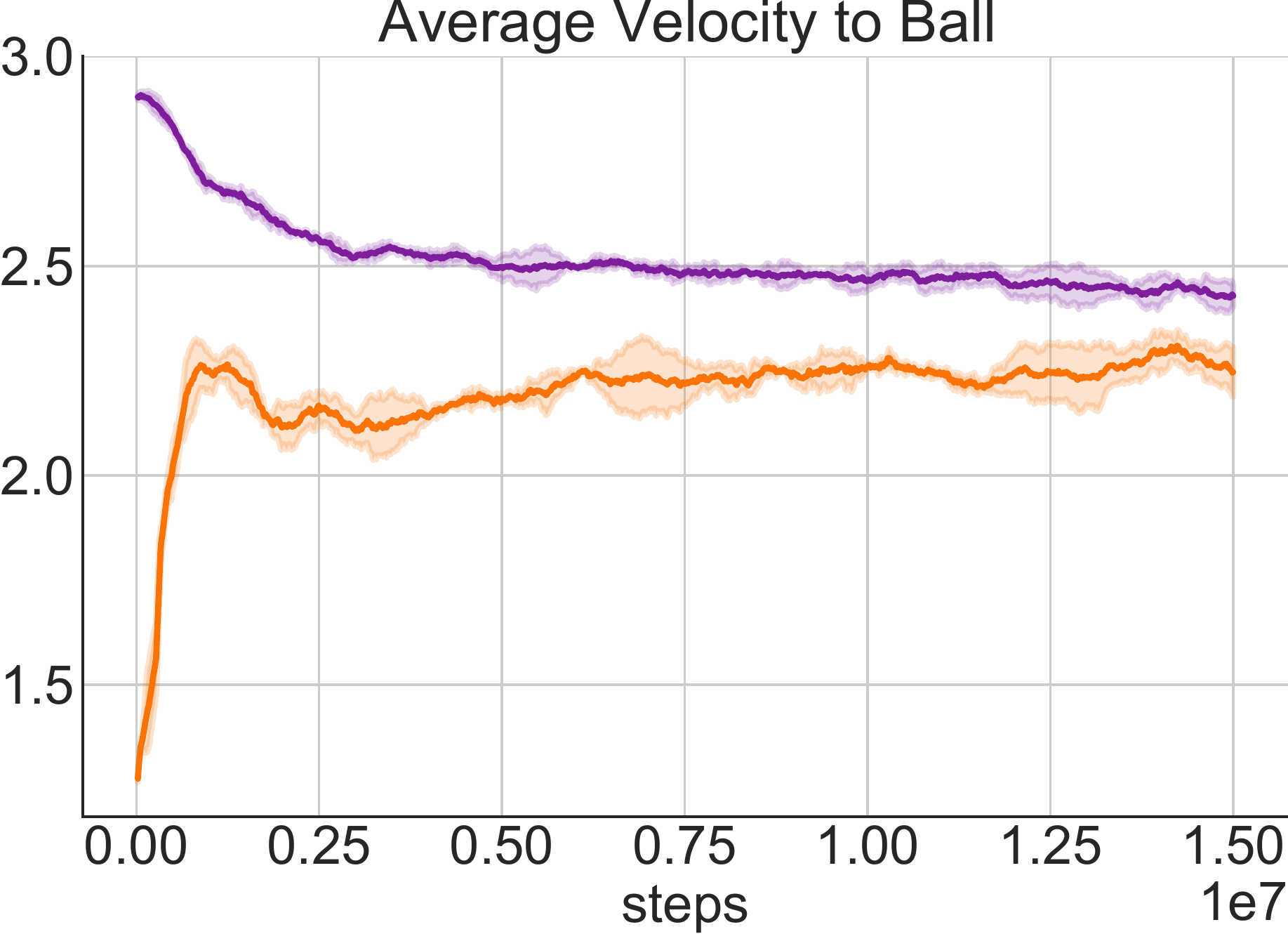} \hfill
    \includegraphics[width=0.24\textwidth]{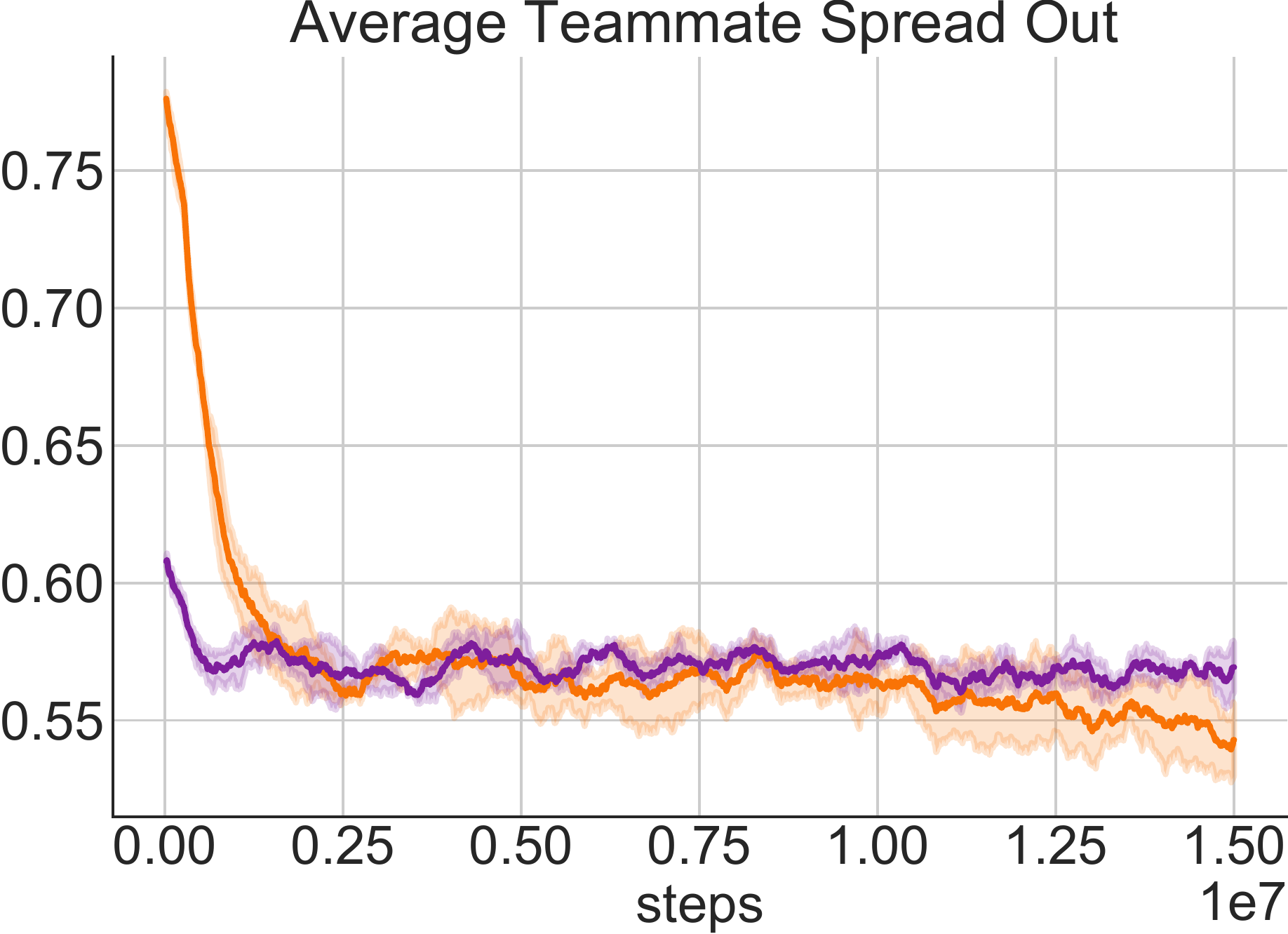} \hfill
    \includegraphics[width=0.24\textwidth]{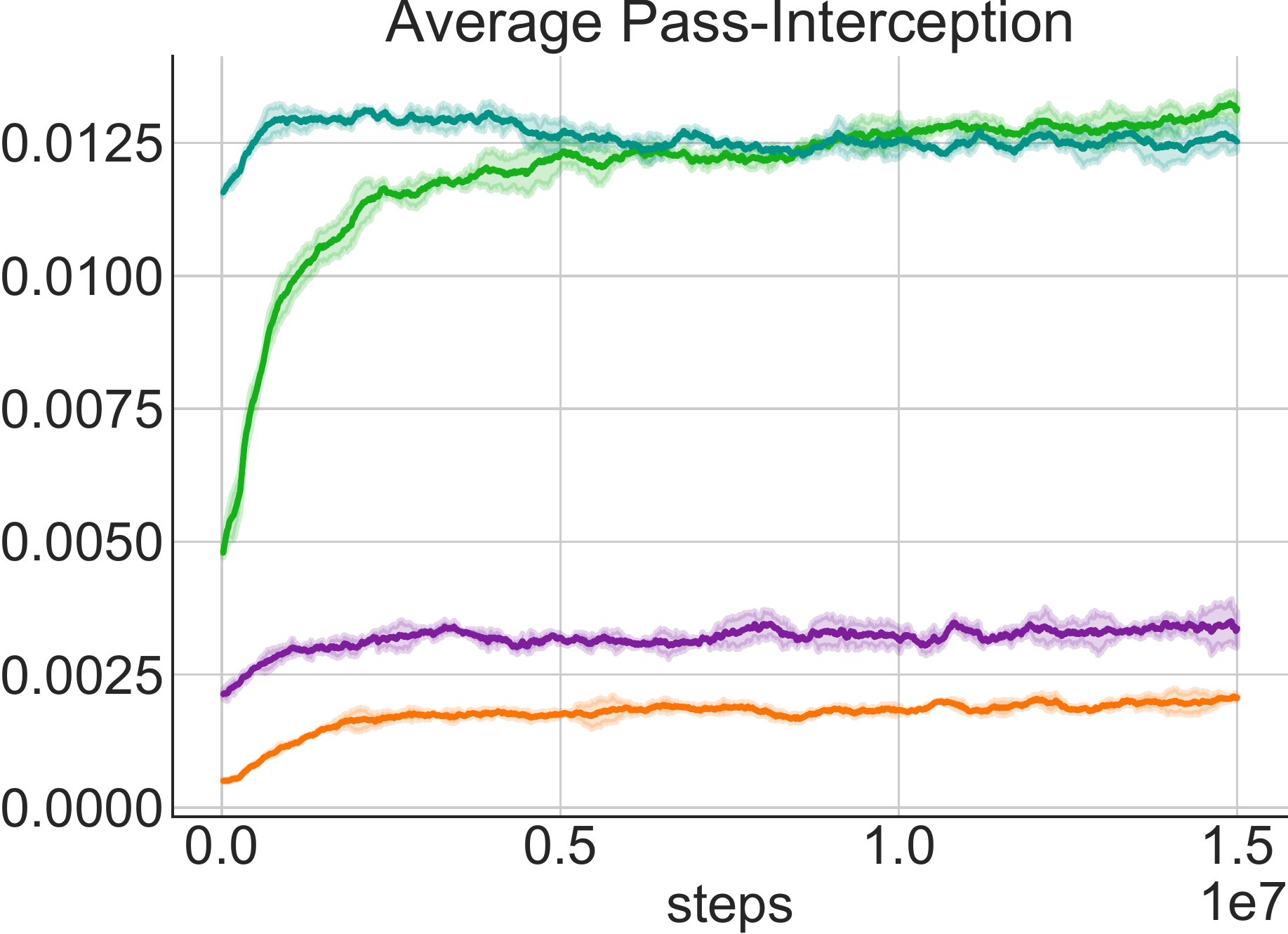} \hfill
    \includegraphics[width=0.24\textwidth]{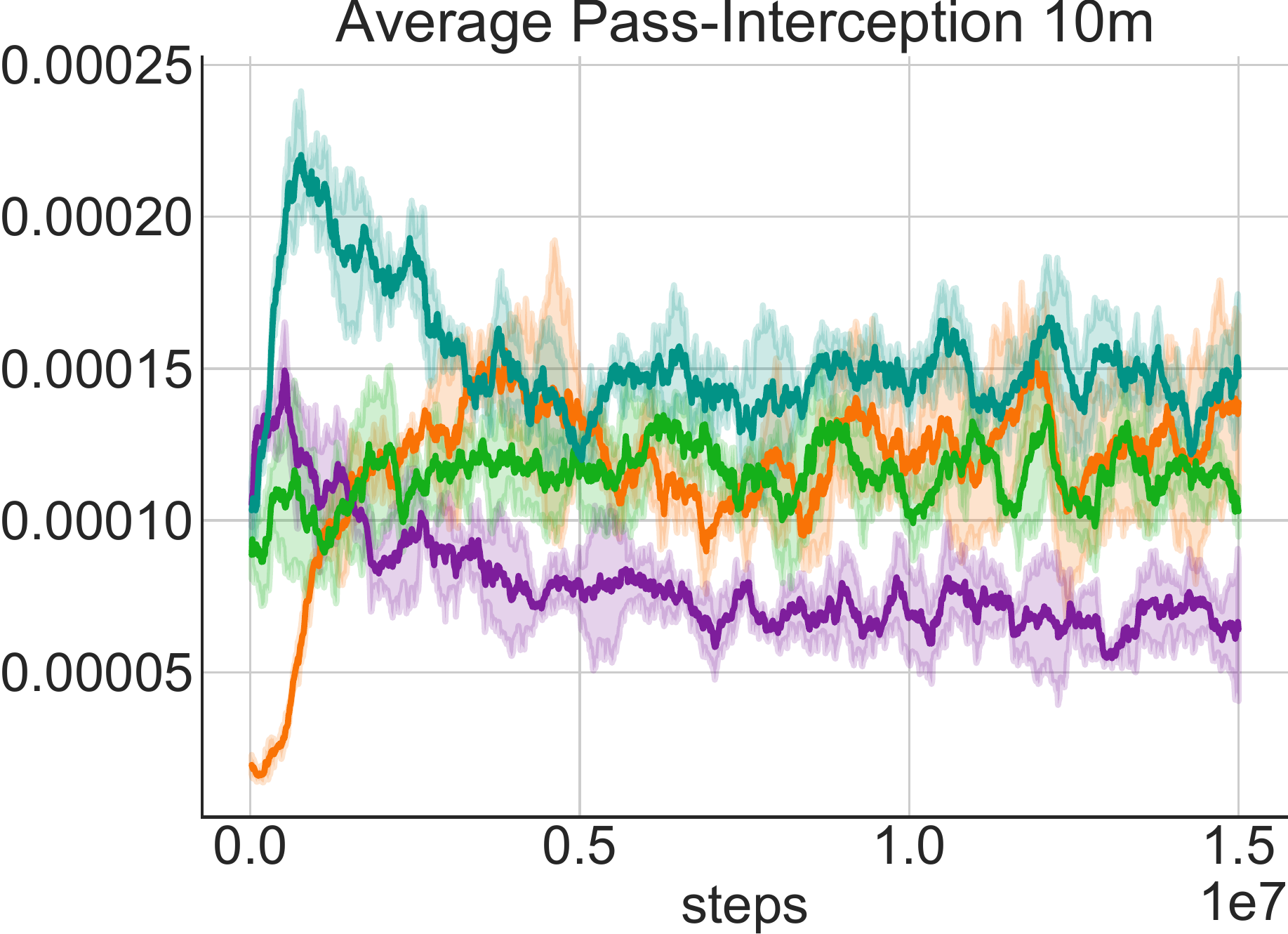}} \vspace{5mm}
    \includegraphics[width=0.75\textwidth]{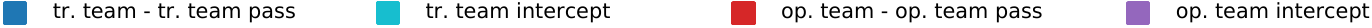}
    \\
    \subfloat[Evolution of performance metrics for team trained in Stage 3 (2v2) under the $\text{DR} + \text{ES} + \text{HCT}$ scheme. \label{fig:metrics-hct}]{%
    \includegraphics[width=0.24\textwidth]{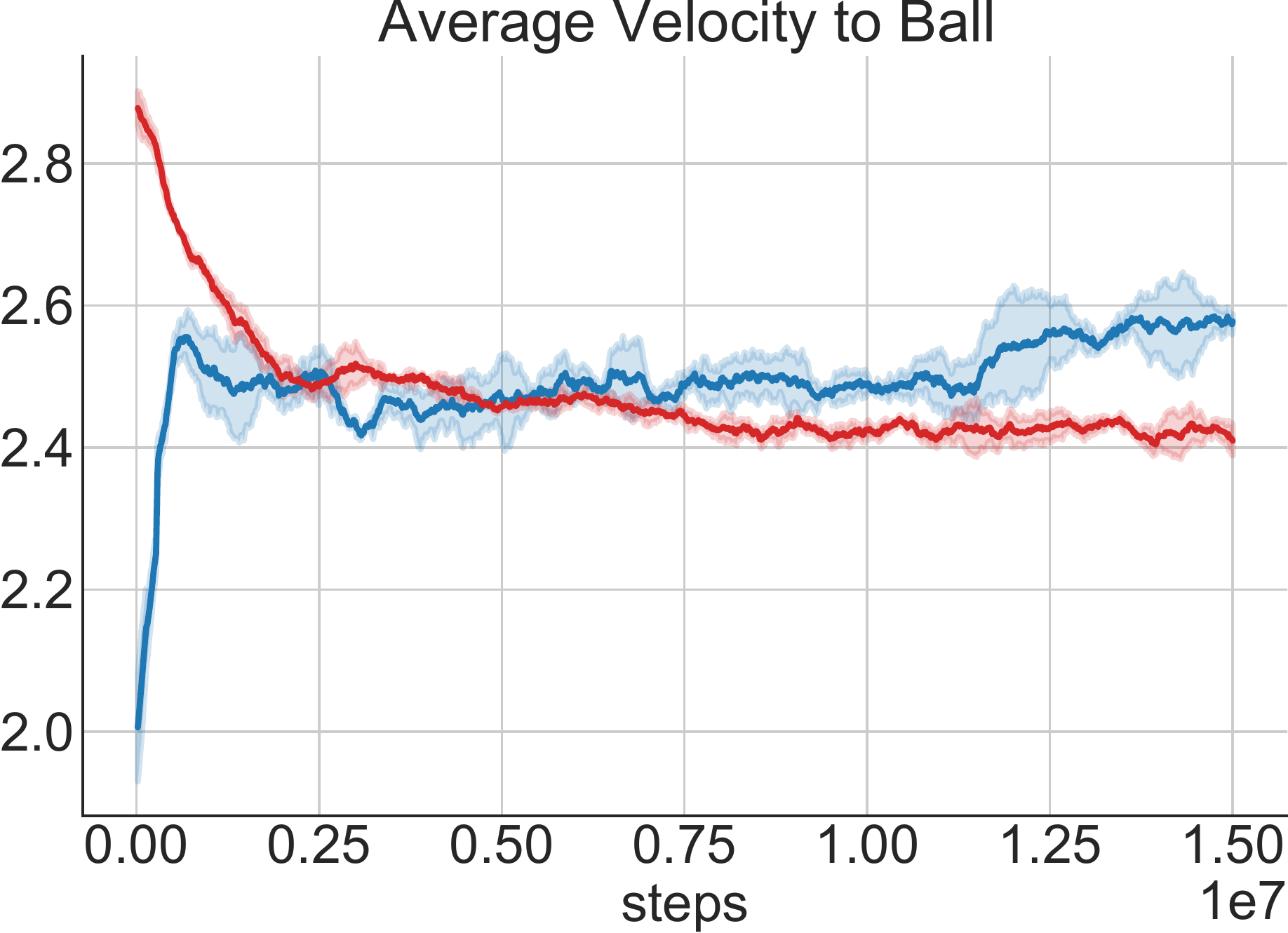} \hfill
    \includegraphics[width=0.24\textwidth]{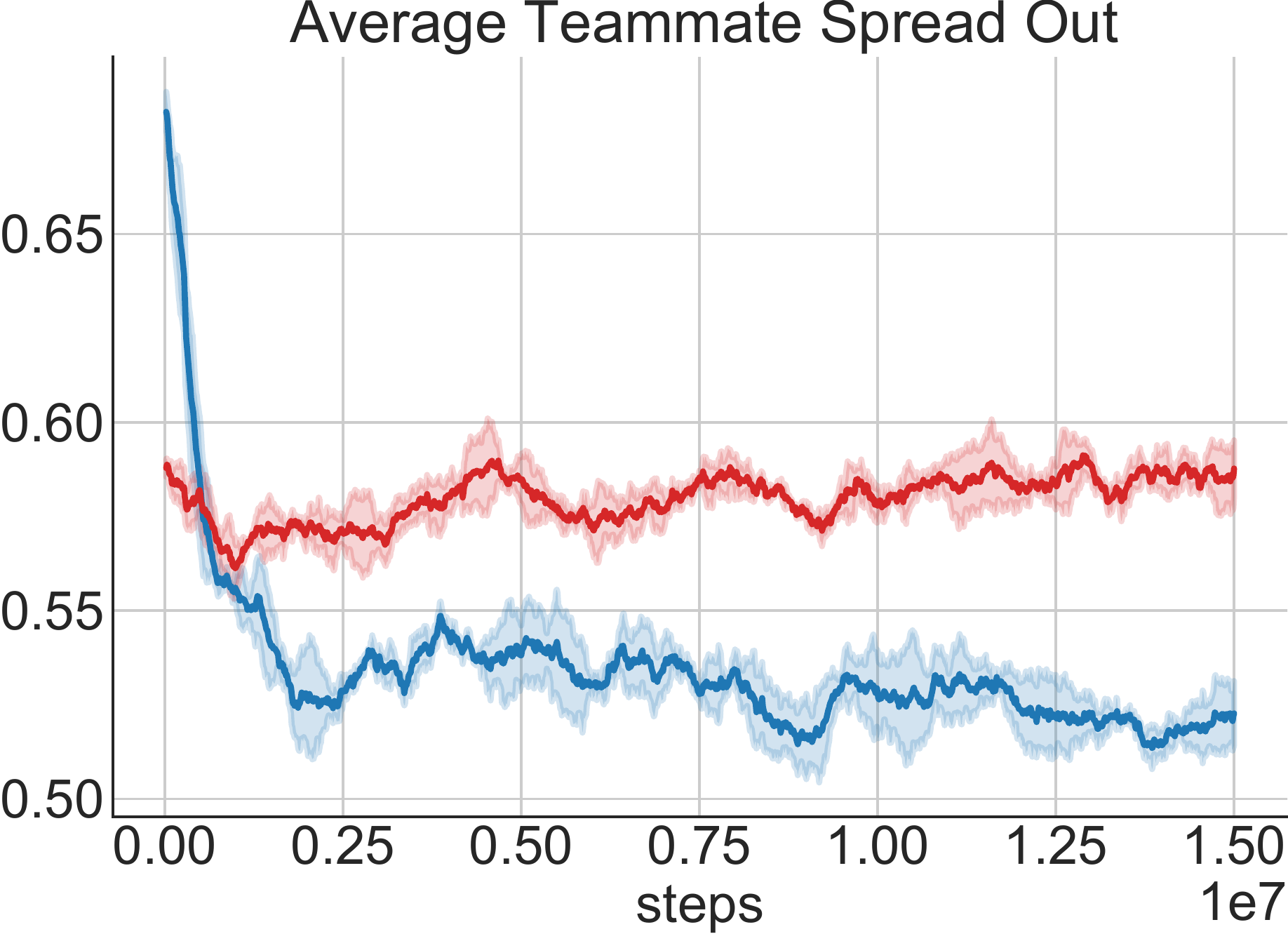} \hfill
    \includegraphics[width=0.24\textwidth]{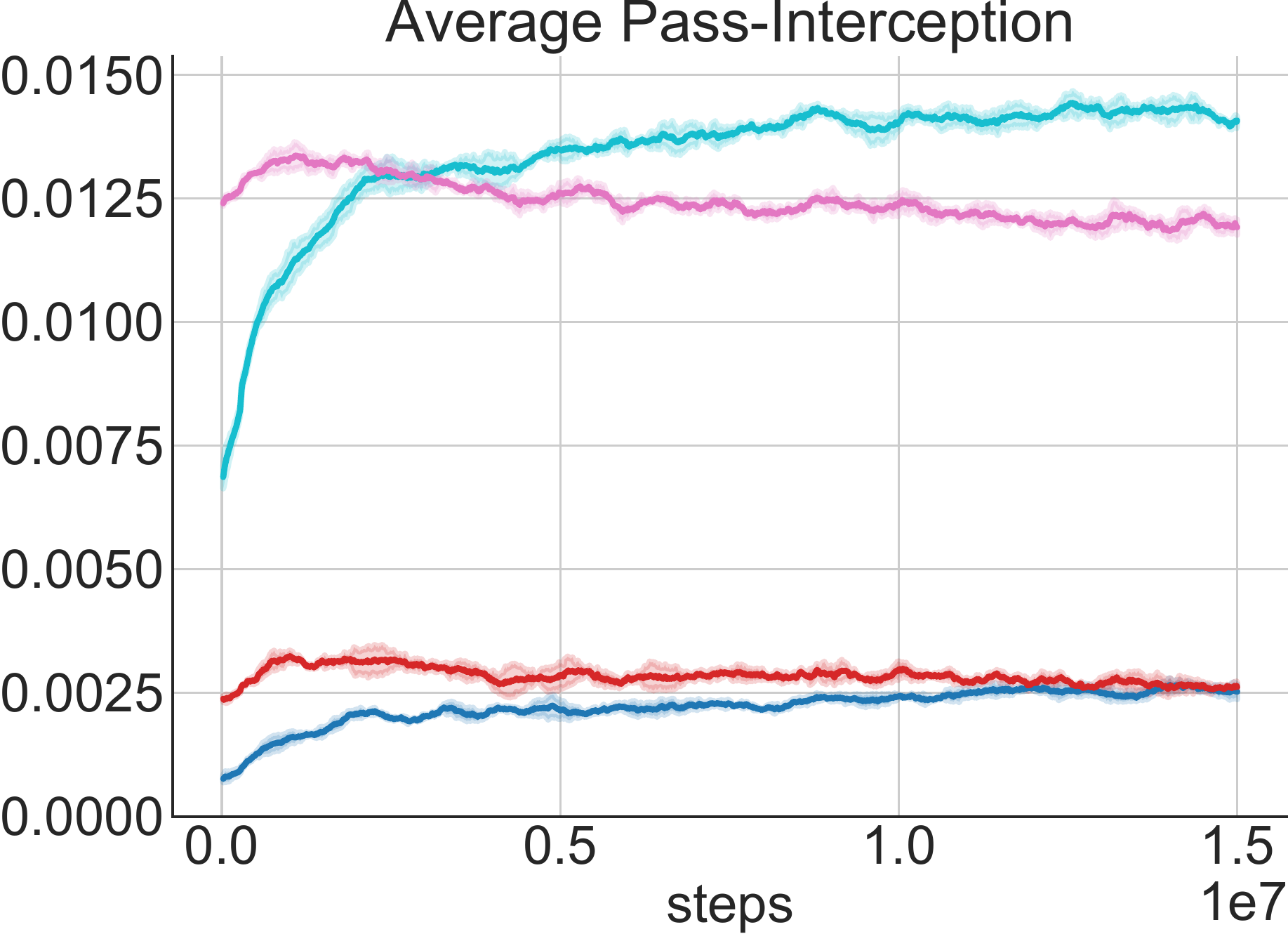} \hfill
    \includegraphics[width=0.24\textwidth]{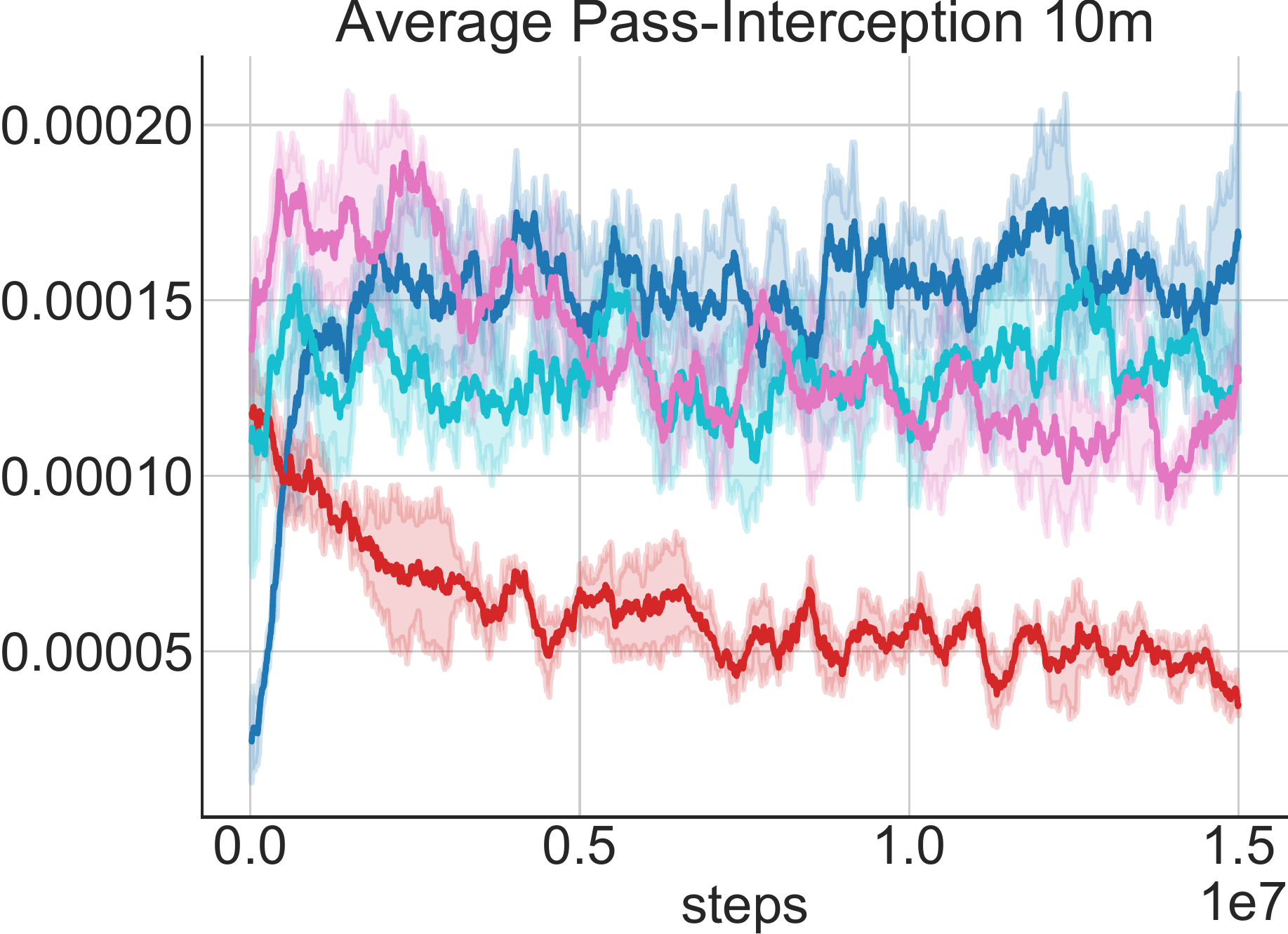}}
    \caption{Evolution of trained (\textit{tr. team}) and opponent team's (\textit{op. team}) performance metrics in stage 3 (2v2), averaged over 3 random seeds.}
    \label{fig:metrics}
\end{figure*}

The control time step defines how often in a given episode the linear acceleration and vertical torque of an agent are controlled. Smaller control time steps allow higher granularity in the control, at the cost of lesser variation between consecutive observations.

In \cite{liu2018emergent} a control time step of 0.05 s is used. We use the same control time step for stages 1, 2, and 3, but also experimented increasing it to 0.1 s for stage 3. Raising the value of this hyper parameter has an effect that is similar to the effect of using frame-skipping \cite{mnih2015human}: sampling transitions while using a higher control time step (HCT), results in a higher variety of experiences, which can increase performance as samples used for training are less correlated. 

As shown in Figure \ref{fig:sr-2v2}, a significant increase of 10\% in the trained team's success rate, measured using the original environment's control time step, can be observed in stage 3 (2v2). 

Additionally, as shown in Figure \ref{fig:payoff-2v2} (which is discussed in more depth in Section \ref{res:agent_selection}), the higher performance in terms of success rate does carry over to better soccer play. This can be seen as the best agents that are trained with a dense reward and a higher control time step, show a high expected goal difference in their favor, when playing against agents trained using the original environment settings.

\subsection{Agent Selection}
\label{res:agent_selection}

Results for the agent selection scheme are shown in Figures \ref{fig:pareto} and \ref{fig:payoff}. Figure \ref{fig:pareto} shows the metrics of the dominant agents per trial in stages 1 and 2. These metrics are obtained by evaluating all resulting agents that were trained for at least 8M steps, on 1,000 episodes of their corresponding task.

For stage 1, only agents trained using the dense reward scheme, which obtain 100\% success rate on the task, are considered. These are evaluated on 1,000 1v0 episodes, and their average \textit{vel-to-ball} and episode length metrics are recorded. These metrics are then used to obtain dominant individuals. This results in 16 dominant agents, as shown in Figure \ref{fig:pareto-1v0}. These 16 agents then compete against each other in a 1v1 setting. Payoff matrices with the expected goal difference among these agents are shown in Figure \ref{fig:payoff-1v0}. As agent N$^\circ$16 has the highest Nash rating, this agent is used as the fixed opponent for stage 2.

Similarly, for stage 2, only agents trained using both, a DR and ES, were considered, as that best performances are obtained when using this scheme (see Figure \ref{fig:sr-1v1}). Agents obtained in the last 20\% steps of the training phase, are then evaluated on the 1v1 task for 1,000 episodes, against the same opponent as in the training phase (the agent N$^\circ 16$ selected from stage 1). The average \textit{vel-to-ball} and episode length metrics were recorded, and agents that did not show top-5\% success rate on the 1v1 task (which translates to $\geq 82.5\%$ success rate), were filtered out. Subsequently, dominant agents per trial, with respect to the recorded metrics, were obtained. Figure \ref{fig:pareto-1v1} shows the average \textit{vel-to-ball} and episode length of the 11 resulting dominant agents. 

Using these top-11 agents, 66 two-agent teams were formed. These teams competed against each other in a 2v2 format. The reduced payoff matrix, that shows the expected goal difference for the 20 agents with top Nash ratings is shown in Figure \ref{fig:payoff-1v1}. As it can be seen, team N$^\circ7$ has the highest Nash rating, so it was used as the fixed opponent for agents trained in stage 3 (2v2).

\subsection{Resulting Behaviors}

Following the approach described in Section \ref{sec:methodology}, policies from stages 1v0, 1v1 and 2v2, are obtained. The various resulting gameplays may be viewed in \url{https://youtu.be/LUruT1A2GOE}.

The following soccer-related skills can be observed when evaluating the trained policies:

\begin{itemize}
    \item \textit{Stage 1 (1v0)}: The agent successfully learns to get close to the ball, and then to kick or dribble the ball towards the goalpost.
    \item \textit{Stage 2 (1v1)}: The agent successfully learns to capture all the skills of the agent trained in stage 1 (1v0), i.e., getting close to the ball, then dribbling or kicking it. Additionally, interesting skills that were observed include feinting, and recovering the ball once in possession of the opponent.
    \item \textit{Stage 3 (2v2)}: In addition to the skills displayed by the agents trained in the previous stage (1v1), the policy obtained in this phase is such that an implicit coordination between teammates is observed. This may be seen by the fact that agents use direct passes and random throw-ins to pass the ball to each other.
\end{itemize}

To quantitatively measure the described behaviors, the same metrics utilized in \cite{liu2018emergent} are considered:

\begin{itemize}
    \item Average velocity to ball: described in Section \ref{subsec:meth_agent_selection}.
    \item Average teammate spread out: measures how often in an episode teammates are more than 5 m away from each other.
    \item Average pass-interception: measures how often in an episode a team passes and intercepts the ball.
    \item Average pass-interception 10 m: same as above, but only passes and interceptions in which the ball has traveled at least 10 m are considered.
\end{itemize}

Figure \ref{fig:metrics} shows the evolution of the performance metrics obtained by some of the teams trained in stage 3, namely, those trained under the schemes $\text{DR} + \text{ES}$ (Fig. \ref{fig:metrics-}) and $\text{DR} + \text{ES} + \text{HCT}$ (Fig. \ref{fig:metrics-hct}). As baselines, metrics obtained by their respective opposing team (formed by agents trained in stage 2 and selected as described in Sect. \ref{res:agent_selection}) are also displayed.

As shown in Figures \ref{fig:metrics-} and \ref{fig:metrics-hct}, an initial rise can be observed when analyzing the \textit{vel-to-ball} metric. This may be attributed to ball chasing behaviors being acquired early on. This metric then sharply drops, to then steadily increase throughout the rest of the training process. This tendency is different from that reported in \cite{liu2018emergent}, where the metric drops throughout the training phase after an initial rise. While this may imply a shift from a predominantly ball chasing behavior to a more cooperative strategy, in our work, a higher \textit{vel-to-ball} metric can observed along with higher \textit{pass} and \textit{pass-10m} metrics, as seen by comparing Figs. \ref{fig:metrics-hct} and \ref{fig:metrics-}.

A similar situation is observed for the \textit{teammate-spread-out} metric. In \cite{liu2018emergent}, this metric rises throughout the training phase (after an initial drop), implying that spread-out teams pass the ball more often as the training progresses. This situation is not observed in our work. On the contrary, we find that higher \textit{teammate-spread-out} values don't correlate with higher pass metrics, as shown in Figs. \ref{fig:metrics-hct} and \ref{fig:metrics-}.

On the other hand, the same tendency reported in \cite{liu2018emergent} of an initially higher \textit{interception-10m} metric, which is later matched by the \textit{pass-10m} metrics, can be observed in both Figs. \ref{fig:metrics-hct} and \ref{fig:metrics-}, however, in the $\text{DR} + \text{ES} + \text{HCT}$ scheme this tendency is more apparent.
 
Finally, it can be seen that the trained team shows higher \textit{pass} and \textit{pass-10m} metrics than the opponent team towards the end of the training process. This is expected, due to the fact that agents that form the opposing teams are trained in stage 2 (1v1), so they are unable to ``observe'' each other.

%% file: conclusion.tex
In this work, we propose a sample efficient method to train a team formed by two agents for playing soccer in the environment introduced in \cite{liu2018emergent}. We use a training curriculum that divides this task in three stages: 1v0, 1v1, and 2v2. The single-agent stage (1v0) is formulated as a classical RL problem, while multi-agent stages (1v1 and 2v2) involve playing against a fixed agent/team, trained in a previous stage. As learning algorithms, we use both vanilla TD3 (for 1v0) and a basic decentralized extension of TD3 for multi-agent RL (for 1v1 and 2v2). In addition, we propose the use of experience sharing, which allows transferring knowledge from previous stages, by leveraging transition tuples experienced by the expert agents.

Results show that coordinated behavior is attainable in a sample-efficient manner, requiring just under 40 M interactions, which represent lesser than 0.1\% of the interactions reported to be needed in the original work \cite{liu2018emergent}. Although the obtained degree of coordination is not as explicit as in \cite{liu2018emergent}, this work shows a new direction which yields promising results, considering the significantly lower training costs.